# Group ICA 2.0: Closing the Gap Between Subjects and Group Latent Decomposition with Copula-Linked Group ICA (CoLiG-ICA)

**Oktay Agcaoglu**[1]
[1] NEFI-Research, LLC
oktay@nefi-research.org

**ABSTRACT:** Group Independent Component Analysis (gICA) is widely used in clinical neuroimaging to decompose high-dimensional functional MRI data into interpretable brain networks. However, conventional gICA is primarily designed to identify components that are shared across subjects. This group-level assumption can limit the recovery of networks present only in individual subjects or subsets of subjects, thereby reducing sensitivity to intersubject heterogeneity in clinical neuroimaging datasets.

We introduce Copula-Linked Group ICA (CoLiG-ICA), a novel algorithm within the Group ICA 2.0 framework that jointly estimates template-linked, cohort-only, and subject-only brain networks within a unified model. The proposed algorithm, CoLiG-ICA, combines ICA-based spatial decomposition, copula-based dependence modeling, and deep learning optimization to retain the consistency and interpretability of template-constrained ICA while enabling additional free components beyond the provided reference networks. By linking subject decompositions to shared templates and jointly estimating cohort-only and subject-only sources, CoLiG-ICA provides a flexible representation of individual variability that is not captured by conventional group priors.

We evaluate CoLiG-ICA using resting-state fMRI data from the UCLA-CNP dataset. We compare the proposed algorithm with conventional constrained ICA to assess estimation of template-linked components, discovery of additional free components, component independence, and the degree to which the estimated components capture subject-level variability beyond the shared group prior.

Compared with MOO-ICAR, CoLiG-ICA showed significantly lower intercomponent spatial dependence, indicating improved subject-level component independence, and significantly reduced motion-related variance in the template-linked components. Additionally, in a schizophrenia-only group analysis, CoLiG-ICA identified three additional resting-state networks beyond the 53 template-linked NeuroMark components: one sensorimotor and two visual networks.

# 1 INTRODUCTION

The human brain is organized through distributed, interacting systems, and functional neuroimaging provides a non-invasive way to measure these systems on a scale. However, functional MRI (fMRI) data are high-dimensional, typically consisting of tens of thousands of voxel-wise time series per subject. This makes dimensionality reduction and latent decomposition central to fMRI analysis, particularly when the goal is to obtain interpretable representations of large-scale brain networks.

Independent Component Analysis (ICA) is one of the most widely used data-driven approaches for fMRI decomposition (Beckmann & Smith, 2004; Bell & Sejnowski, 1995; Hyvärinen & Oja, 2000; McKeown & Sejnowski, 1998). In contrast to model-based approaches that require a predefined task design or temporal response model, ICA estimates latent sources directly from the data by seeking components that are statistically independent. Under standard linear ICA assumptions, the latent sources are identifiable up to scale and permutation ambiguities, which provides a well-defined decomposition target and supports interpretable representation learning.

ICA was introduced to fMRI in the late 1990s and has since become a standard tool for exploratory functional neuroimaging analysis. In fMRI, ICA can be formulated as either spatial ICA, which estimates statistically independent spatial maps and their associated time courses, or temporal ICA (Calhoun et al., 2001b), which estimates independent temporal patterns. Because fMRI datasets generally contain far more voxels than time points, spatial ICA has been more widely adopted than temporal ICA. Although gICA is applicable to both task-based and resting-state fMRI, it has been particularly widely used in resting-state studies, where the resulting components are commonly interpreted as intrinsic functional brain networks, often referred to as resting-state networks (RSNs) (Calhoun et al., 2009; Damoiseaux et al., 2006). Resting-state fMRI is also more widely available than task-based fMRI and is less affected by variability in task compliance. Accordingly, this paper focuses on applying spatial gICA to resting-state fMRI data.

Early applications of ICA to fMRI were typically performed at the individual-subject level (Beckmann & Smith, 2004; McKeown & Sejnowski, 1998), decomposing each subject's data into spatial components and associated time courses. Although this approach allowed subject-specific network estimation, it created a practical barrier for group analysis: components estimated independently for each subject had to be matched, ordered, and interpreted across individuals. This was especially challenging in resting-state fMRI, where components may reflect both neural networks and artifacts, and manual inspection can introduce additional effort and variability.

Group ICA (gICA) was introduced to address this limitation by estimating a common set of group-level components from multi-subject fMRI data (Calhoun et al., 2001a, 2009). In the commonly used temporal-concatenation framework, subject-level data are first reduced, concatenated across subjects, and then decomposed using ICA to estimate spatial components shared across the cohort. Subject-specific spatial maps and time courses are then obtained through back-reconstruction or related projection/regression procedures. gICA is illustrated in Figure 1. This strategy greatly simplifies cross-subject comparison because components are aligned by construction, enabling group-level inference and downstream statistical analysis.

However, this improvement comes at the cost of reduced sensitivity to subject-specific variability (Allen et al., 2012; Calhoun et al., 2009). Group ICA has two important limitations, one practical and one theoretical. From a practical perspective, gICA estimates a decomposition tailored to a particular cohort (Calhoun et al., 2009; Du et al., 2020; Varoquaux et al., 2010). When a different cohort or dataset is analyzed, the decomposition often must be re-estimated, and the resulting components may differ across studies, including differences in both intrinsic networks and artifact components. These cohort-specific decompositions make between-group comparisons challenging and can reduce consistency across studies. Moreover, component labeling often requires manual inspection (Agcaoglu et al., 2019; Allen et al., 2011; Griffanti et al., 2017; Rashid et al., 2018), which introduces subjectivity; different researchers may assign similar components to different functional domains or classify artifacts differently. From a theoretical perspective, ICA is valued because it exploits higher-order statistical structure (Bell & Sejnowski, 1995; Hyvärinen & Oja, 2000). In standard gICA, however, this higher-order information is used only after subject-level dimensionality reduction and group-level aggregation, rather than being directly leveraged at the subject level (Calhoun et al., 2009; Erhardt et al., 2011).

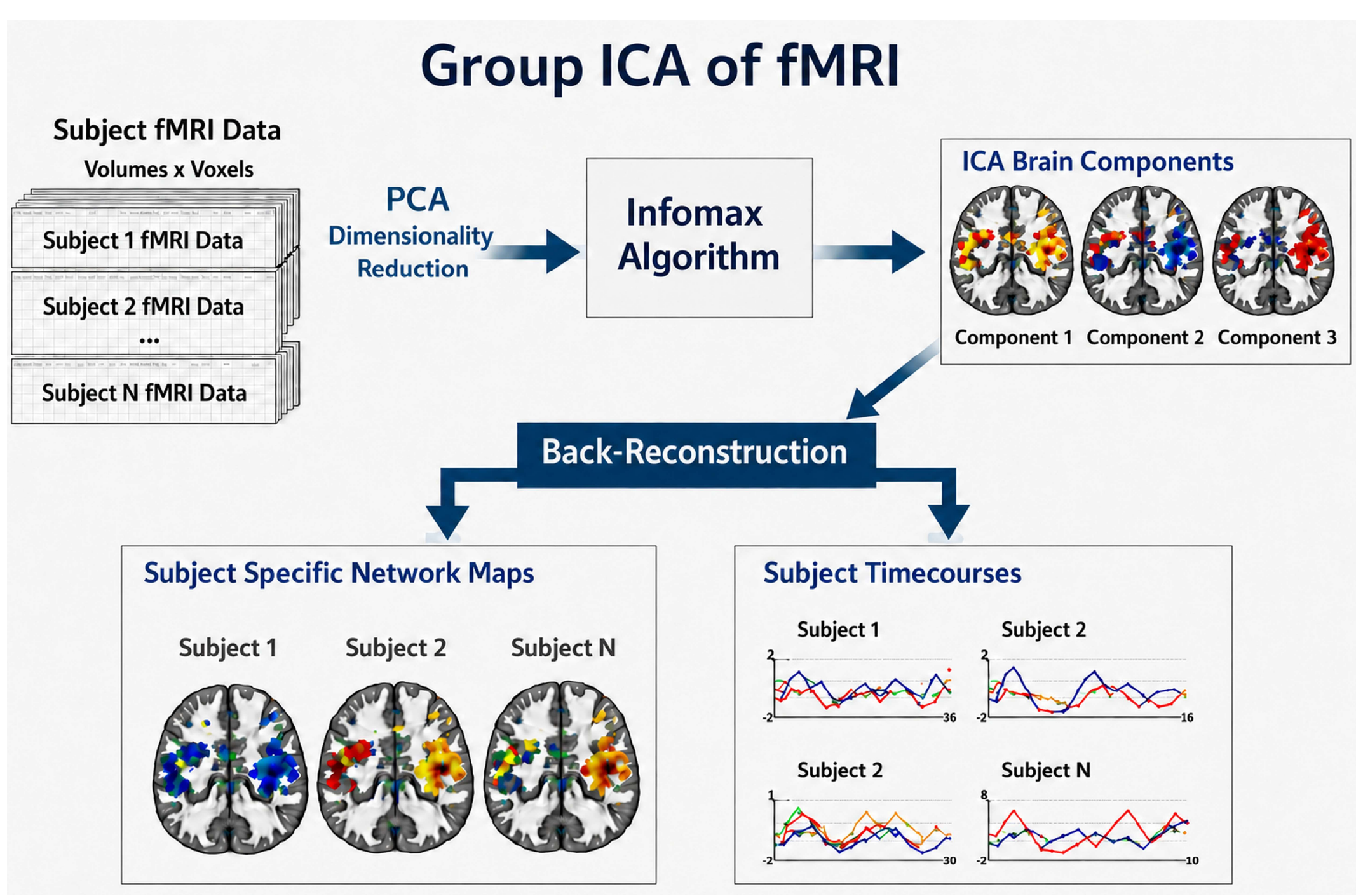

**Figure 1: Conventional group ICA framework for fMRI. Subject fMRI datasets are concatenated and reduced using PCA before group-level ICA estimates common spatial components. Back-reconstruction is then used to obtain subject-specific spatial maps and associated time courses.**

There are alternative approaches to the subject component matching problem besides gICA. For example, Independent Vector Analysis (IVA) estimates subject-level decompositions while jointly encouraging alignment across subjects (Kim et al., 2006; Lee et al., 2008; Michael et al., 2014). By modeling statistical dependence across datasets, IVA can preserve subject-level variability while linking corresponding components across individuals. However, IVA is a multiset optimization problem and can become computationally and memory intensive as the number of subjects increases (Vu et al., 2024). In practice, published IVA applications to multi-subject resting-state fMRI remain limited in scale; for example, a recent constrained IVA study analyzed resting-state fMRI data from 98 subjects and described this as the largest number of subjects used by IVA algorithms (Vu et al., 2024).

To address these issues, constrained ICA approaches were proposed. These approaches leverage higher-order statistics while decomposing each subject using a common ICA template or reference, thereby promoting consistency of components across subjects (Du et al., 2020; Du & Fan, 2013). For example, group-information-guided ICA estimates subject-specific independent components using group-level information as guidance, and NeuroMark extends this idea into an automated, adaptive pipeline for identifying reproducible fMRI markers across datasets, studies, and disorders (Du et al., 2020; Du & Fan, 2013). However, these approaches are also limited: they assume that a fixed template provides an adequate representation for each subject, even though clinical populations may differ in the number, structure, and prominence of functional networks. They also do not explicitly allow additional subject-only or cohort-only networks beyond the provided template. This matters both for mitigating subject- or cohort-specific artifacts and for capturing meaningful resting-state networks that may be specific to a subgroup or individual.

Group ICA 2.0 is a broader methodological framework that combines deep learning optimization and copula-based dependence modeling to address key limitations of conventional gICA and enable flexible multimodal linkage and group- and subject-level decompositions. CLiP-ICA (Agcaoglu, Alacam, et al., 2024) was the first method developed within this framework. In this study, we address a longstanding gap in group-level decomposition by introducing Copula-Linked Group ICA (CoLiG-ICA), which was initially presented in a preliminary conference abstract (Agcaoglu et al., 2025a). The present work provides the full methodological formulation and an expanded empirical evaluation.

CoLiG-ICA retains the benefits of template-constrained network estimation (Du et al., 2020; Du & Fan, 2013; Vu et al., 2024) while allowing cohort-specific and subject-specific brain networks beyond the provided template. CoLiG-ICA integrates ICA, copula-based dependence modeling, and deep learning optimization to estimate flexible group and individual decompositions.

CoLiG-ICA estimates cohort-specific and subject-specific brain networks in addition to the provided template-linked networks while maximizing independence among all components. It also allows subjects to have different model orders, enabling multi-scale analysis across individuals and cohorts.

## 2 METHOD

### 2.1 Overview: Copula-Linked Group ICA (CoLiG-ICA)

CoLiG-ICA is a template-guided, scalable latent decomposition algorithm designed to retain the practical advantages of constrained/template ICA (consistent component identity across subjects) while closing the gap between group-level and subject-level decomposition. The proposed algorithm, CoLiG-ICA, couples each subject's estimated functional networks to a provided template set (e.g., NeuroMark-53 or a study-specific gICA template), while also allowing additional cohort-only and subject-only components. In contrast to standard gICA (where higher-order statistics are primarily leveraged after group-level reduction), CoLiG-ICA applies ICA-style higher-order modeling at the subject level and uses copula-based dependence modeling to flexibly enforce template coupling without over-constraining the marginal distributions.

At a high level, CoLiG-ICA optimizes a joint objective that (i) encourages statistical independence among all recovered components (template + free components) and (ii) enforces template alignment via a copula dependence term between each template map and its matched subject map. The "free" components are not matched to the template, enabling discovery of cohort-only and subject-only networks (including subgroup-specific networks and subject-specific artifacts). Figure 2 presents a block diagram of CoLiG-ICA.

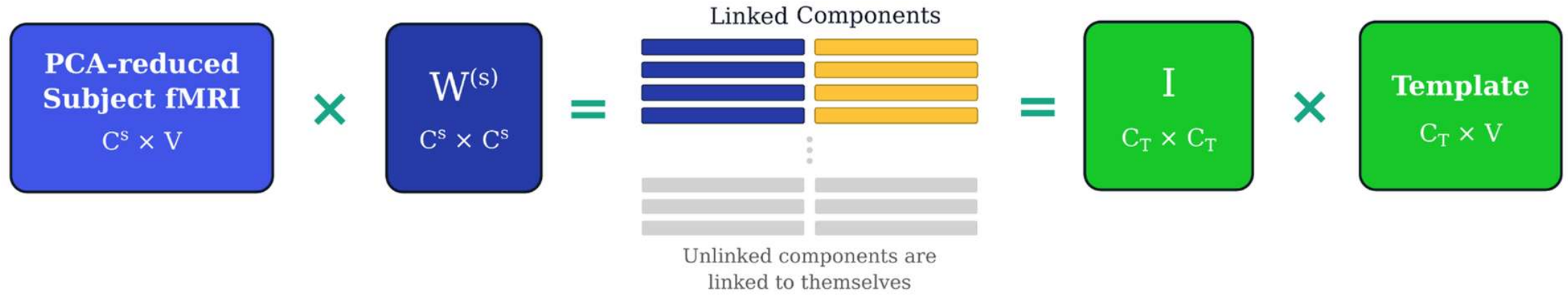

**Figure 2: Block diagram of CoLiG-ICA. Template-linked components are coupled with the reference maps using Gaussian copulas, while free components remain unconstrained. The subject-specific unmixing matrix $W^s$ is estimated using gradient-based optimization within a deep learning framework.**

### 2.2 Data model and notation (spatial ICA)

For subject $s$, let $X^s \in \mathbb{R}^{T\times V}$ denote the preprocessed rs-fMRI data, with $T$ time points and $V$ voxels. Spatial ICA models the data as $X^s = A^s M^s$, where $A^s \in \mathbb{R}^{T\times C^s}$ contains the component time courses and $M^s \in \mathbb{R}^{C^s\times V}$contains the spatial maps. CoLiG-ICA partitions the model order as $C^s = C_T + C_F^s$, where $C_T$ components are template-linked and $C_F^s$ components are free of template constraints.

Subject-level PCA projection $P^s$ reduces and whitens the data: $Z^s = P^s X^s, M^s = W^s Z^s$, where $W^s \in \mathbb{R}^{C^s\times C^s}$ is the subject-specific ICA unmixing matrix. Thus, $M^s = U^s X^s$, where $U^s = W^s P^s$. Unlike conventional group ICA, CoLiG-ICA optimizes $W^s$ separately for each subject.

### 2.3 Template-linked and free component decomposition

For subject $s$, let $Z^s \in \mathbb{R}^{C^s \times V}$ denote the dimension-reduced fMRI data and $M^s = W^s Z^s$ denote the estimated component maps, where $W^s \in \mathbb{R}^{C^s \times C^s}$. The components are partitioned as $C^s = C_T + C_F^s$, where $C_T$ components are linked to the template and $C_F^s$ are free components learned without template constraints.

Let $\mathcal{T} = \{t_i\}_{i=1}^{C_T}$ denote the set of template maps. Using Sklar's theorem (Sklar, 1959), the likelihood, up to terms independent of $W^s$, is

$$p(Z^s, \mathcal{T} \mid W^s) \propto \mid \det W^s \mid^V \prod_{v=1}^{V} \left[ \prod_{i=1}^{C^s} p_i\,(m_{iv}^s) \prod_{i=1}^{C_T} c_{\rho_i}\,(u_{iv}^s, \tilde{u}_{iv}) \right],$$

where $u_{iv}^s = F_i(m_{iv}^s)$, $\tilde{u}_{iv} = F_{T_i}(t_{iv})$, and $c_{\rho_i}$ is the copula density linking subject component $i$ to its matched template component. For the Gaussian copula,

$$c_{\rho_i}(u, \tilde{u}) = \frac{1}{\sqrt{1 - \rho_i^2}} \exp\left[ -\frac{1}{2} \mathbf{q}^{\mathsf{T}} (R_i^{-1} - I) \mathbf{q} \right],$$

where

$$\mathbf{q} = \begin{bmatrix} \Phi^{-1}(u) \\ \Phi^{-1}(\tilde{u}) \end{bmatrix}, R_i = \begin{bmatrix} 1 & \rho_i \\ \rho_i & 1 \end{bmatrix}, \mid \rho_i \mid < 1.$$

The model is optimized by minimizing the negative log-likelihood. Only the first $C_T$ components receive copula constraints. The remaining $C_F^s$ components have no template-coupling term and are therefore free to capture subject-specific or cohort-enriched patterns not represented in the template. Component-template matching is performed during initialization and may be updated during early optimization until the assignments stabilize.

## 3 ANALYSIS

### 3.1 Dataset

We used the publicly available UCLA Consortium for Neuropsychiatric Phenomics (UCLA-CNP) dataset (Poldrack et al., 2016). The first eight functional volumes were discarded to allow the MR signal to reach steady state. Resting-state fMRI data were then preprocessed using a standard SPM12 pipeline (https://www.fil.ion.ucl.ac.uk/spm/), including slice-timing correction, rigid-body realignment, coregistration to the T1-weighted image, tissue segmentation, normalization to Montreal Neurological Institute (Mazziotta et al., 2001) space at 3 × 3 × 3 mm resolution, and spatial smoothing with a 6-mm full-width at half-maximum Gaussian kernel. Head motion was assessed using the six SPM realignment parameters and framewise displacement.

Scans were flagged during quality control if mean framewise displacement exceeded 0.25 mm, maximum translation exceeded 3 mm, maximum rotation exceeded 3°, or fewer than 144 functional volumes remained. After quality control, 190 participants were retained, including 99 healthy controls (HC), 30 participants with attention-deficit/hyperactivity (ADHD) disorder, 33 with bipolar disorder, and 28 with schizophrenia (SCHZ). The resting-state scans had a repetition time of 2 seconds.

### 3.2 ICA Analysis

We applied CoLiG-ICA and multi-objective optimization ICA with reference (Du et al., 2020; Du & Fan, 2013) to each subject for comparison. Before decomposition, the mean across all time points was removed from the fMRI time series at each voxel for both methods. The NeuroMark template, containing 53 RSNs, was used as the reference template. To provide a fair comparison, we selected a model order of 100 for CoLiG-ICA, matching the model order used in the original study that generated the NeuroMark template. This configuration yielded 53 template-linked components and 47 additional free components. In contrast, MOO-ICAR is restricted by design to the 53 components specified by the reference template and does not estimate additional components outside that template. For functional network connectivity analysis, the six rigid-body realignment parameter time courses were regressed from the estimated component time courses. The residual time courses were then detrended, despiked, and band-pass filtered between 0.01 and 0.1 Hz. Motion-related $R^2$ was evaluated separately using component time courses before motion regression. We compared the spatial dependencies of the estimated component maps, calculated functional network connectivity matrices, and evaluated group differences across the diagnostic groups and healthy controls.

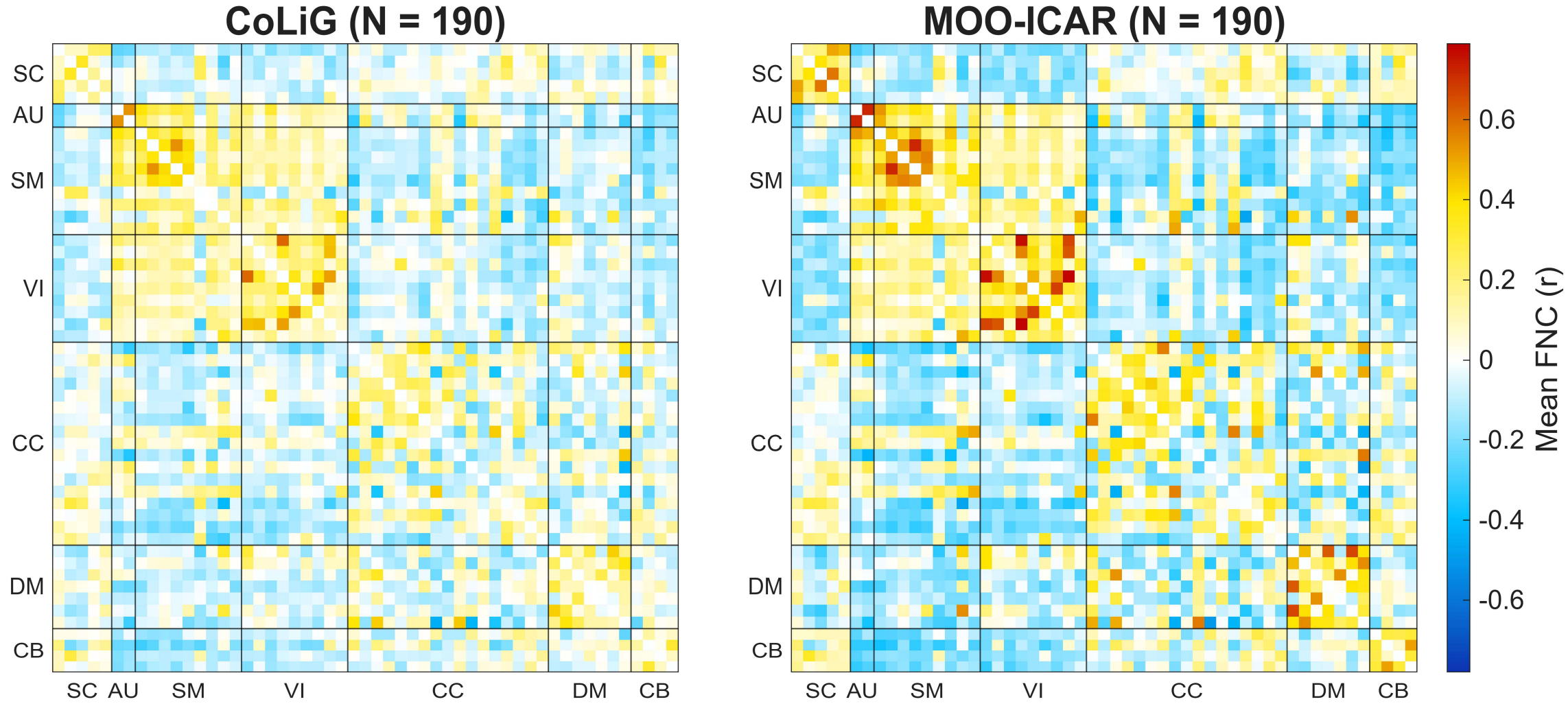

**Figure 3: Mean functional network connectivity (FNC) matrices obtained using CoLiG-ICA and MOO-ICAR. The functional domains are subcortical (SC), auditory (AU), sensorimotor (SM), visual (VI), cognitive control (CC), default mode (DM), and cerebellar (CB).**

### 3.3 Cohort-Only Group ICA Analysis

We also applied CoLiG-ICA at the group level to identify additional cohort-specific components. For this analysis, we used data from participants with schizophrenia. We performed two-level PCA, reducing the data first to 120 and then to 100 dimensions. CoLiG-ICA was then applied to estimate 53 template-linked and 47 free components.

## 4 RESULTS

CoLiG-ICA revealed a clear modular organization in the FNC matrices. Its mean FNC matrix was broadly consistent with that of MOO-ICAR (Figure 3), showing positive within-domain connectivity and similar positive and negative between-domain patterns. CoLiG-ICA produced lower absolute connectivity for several network pairs. This may suggest that the free components captured residual shared variance, including nuisance-related effects, while preserving the principal functional organization.

We next assessed motion-related variance in the template-linked component time courses before nuisance regression. The component and motion time courses were detrended and band-pass filtered using the same preprocessing procedures. Motion-related variance was quantified using the coefficient of determination, $R^2$, for both the six-parameter rigid-body motion model and the Friston 24-parameter model. CoLiG-ICA showed significantly lower motion-related $R^2$ values than MOO-ICAR under both models, with mean reductions of 0.0199 ($p = 3.3 \times 10^{-37}$) and 0.0198 ($p =$ $6.6 \times 10^{-45}$), respectively (Figure 4). These findings are consistent with the additional free components capturing motion-related variance that would otherwise remain in the template-linked components.

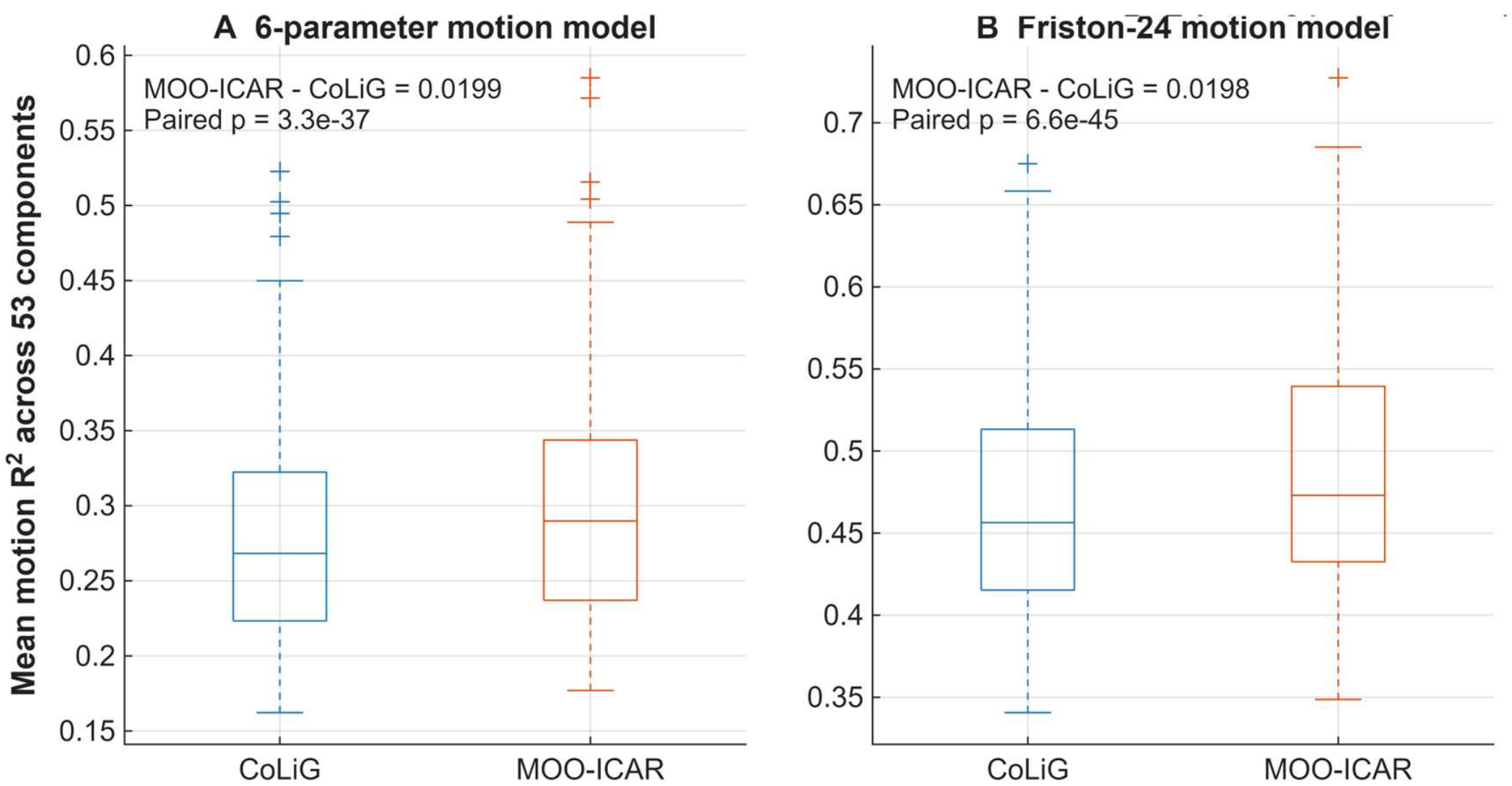

**Figure 4: Motion-related variance. Mean $R^2$ across the 53 template-linked components for the six-parameter and Friston 24-parameter motion models.**

We also evaluated spatial dependence among the 53 template-linked subject-specific component maps. For each subject, spatial dependence was summarized as the mean absolute off-diagonal pairwise spatial correlation (Figure 5). CoLiG-ICA showed substantially lower spatial dependence than MOO-ICAR, with a mean paired difference of 0.06463 (MOO-ICAR minus CoLiG-ICA). This reduction was highly significant across the 190 subjects (paired $t$-test: $p = 1.53 \times 10^{-2}$ , $d_z = 13.075$) and was confirmed by a Wilcoxon signed-rank test ($p = 4.29 \times 10^{-33}$). These findings indicate that CoLiG-ICA more effectively maximized subject-level spatial independence.

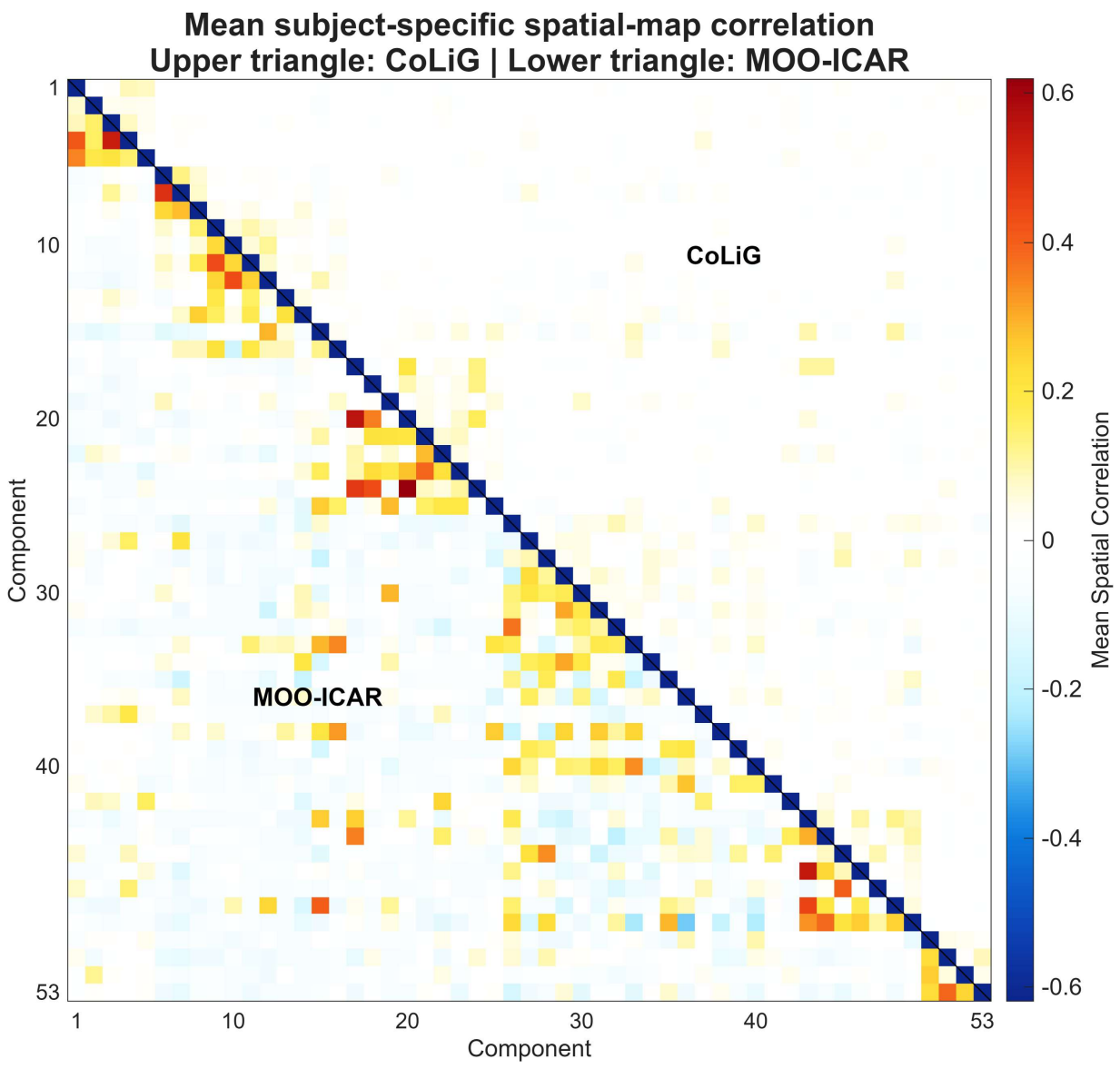

**Figure 5: Mean pairwise spatial correlations among the 53 subject-specific component maps for CoLiG-ICA and MOO-ICAR.**

Finally, we evaluated diagnostic-group differences in functional network connectivity. Age, sex, site, and mean framewise displacement were included as covariates in a linear regression model, and the effect of diagnosis was tested for each connectivity measure. After FDR correction across 1,378 connectivity measures (q<0.05), neither CoLiG-ICA nor MOO-ICAR identified significant diagnostic-group differences. The relatively small diagnostic subgroups may have limited statistical power to detect subtle effects. By contrast, a previous analysis of FBIRN schizophrenia data found that CoLiG-ICA identified more significant group differences after FDR correction than MOO-ICAR (Agcaoglu et al., 2025a).

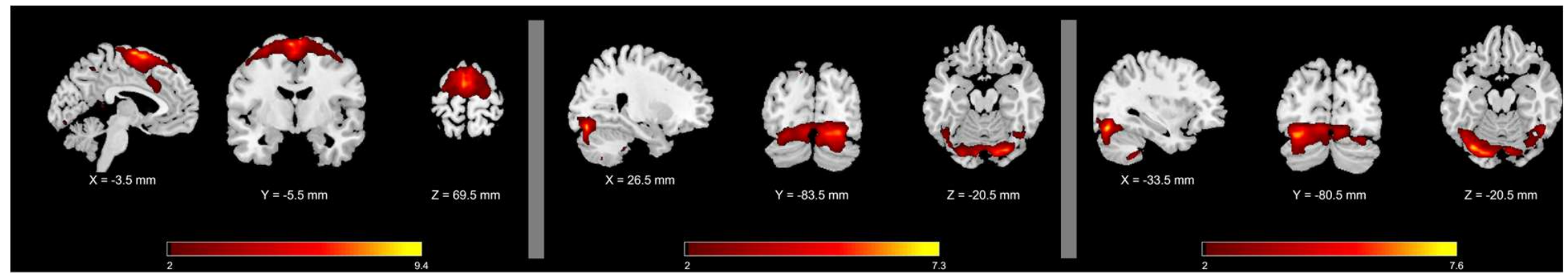

**Figure 6: Three cohort-specific components identified in the schizophrenia cohort in addition to the template-linked components. The left component was identified as a sensorimotor network, whereas the middle and right components were identified as visual networks.**

Accordingly, the cohort-only analysis identified three additional components beyond the 53 NeuroMark components. The 47 free components were visually evaluated for spatial coherence, gray-matter localization, and minimal ventricular, edge, and white-matter involvement. Three components meeting these criteria were classified as plausible resting-state networks based on their anatomical distributions. One was classified as a sensorimotor network, while the other two were classified as visual networks. The visual components formed a lateralized pair, with one predominantly involving the left hemisphere and the other predominantly involving the right hemisphere. These components are presented in Figure 6. Spectra were calculated separately for each participant and then averaged across participants at each frequency. As shown in Figure 7, the spectra exhibited characteristics consistent with resting-state networks, including greater power at lower frequencies and reduced power at higher frequencies (Allen et al., 2011). The displayed dynamic range represents the difference between the peak spectral power and the minimum power at frequencies above the peak, while the low-to-high-frequency power ratio was calculated using predefined frequency ranges of Group ICA Of fMRI Toolbox (GIFT) (https://github.com/trendscenter/gift).

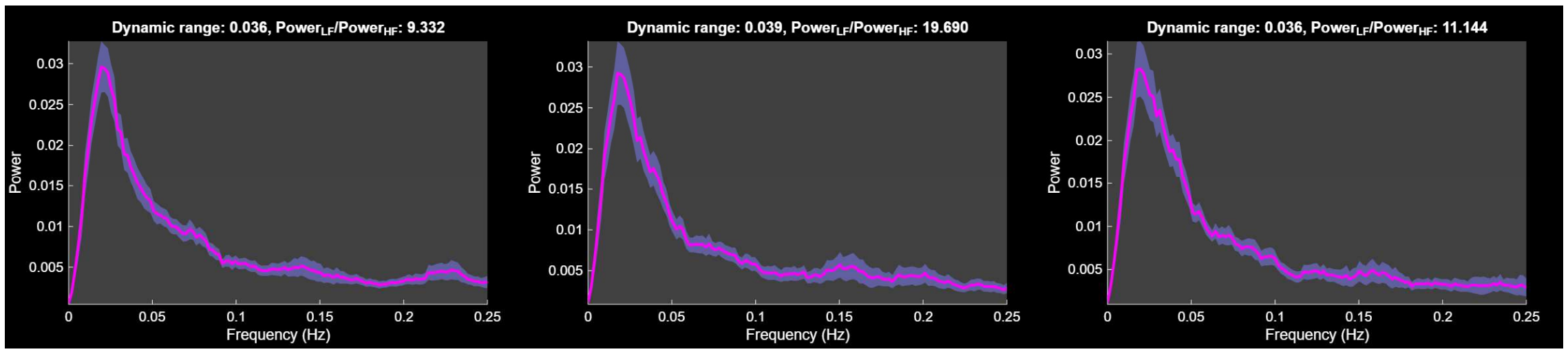

**Figure 7: Average power spectra of the three components, displayed from left to right in the same order as in Figure 6. The magenta curves represent the mean power spectra across participants, and the gray shaded regions represent the mean ± one standard error of the mean (SEM). The spectra show greater power at lower frequencies and reduced power at higher frequencies, consistent with the spectral characteristics of resting-state networks. The dynamic range and low-to-high-frequency power ratio are reported above each spectrum.**

We also calculated the FNC profiles of these components and displayed them within their corresponding functional domains in Figure 8. Their connectivity patterns aligned well with the existing modular organization of the FNC matrix.

Preliminary visual inspection of the subject-level free components in this initial study suggested a mixture of nuisance-related and potentially meaningful network patterns. Although these components were not systematically evaluated, their benefits were reflected in significant reductions in motion-related variance and intercomponent spatial dependence among the NeuroMark-linked components.

Future work will systematically evaluate the free components and investigate additional constraints, such as multimodal coupling, to improve the recovery and interpretability of meaningful subject-specific networks. This direction is motivated by findings from CLiP-ICA (Agcaoglu, Alacam, et al., 2024; Agcaoglu et al., 2025b; Agcaoglu, Silva, et al., 2024), which demonstrated benefits for RSN estimation and artifact mitigation by coupling fMRI with structural MRI and using structural information to guide latent-space regularization.

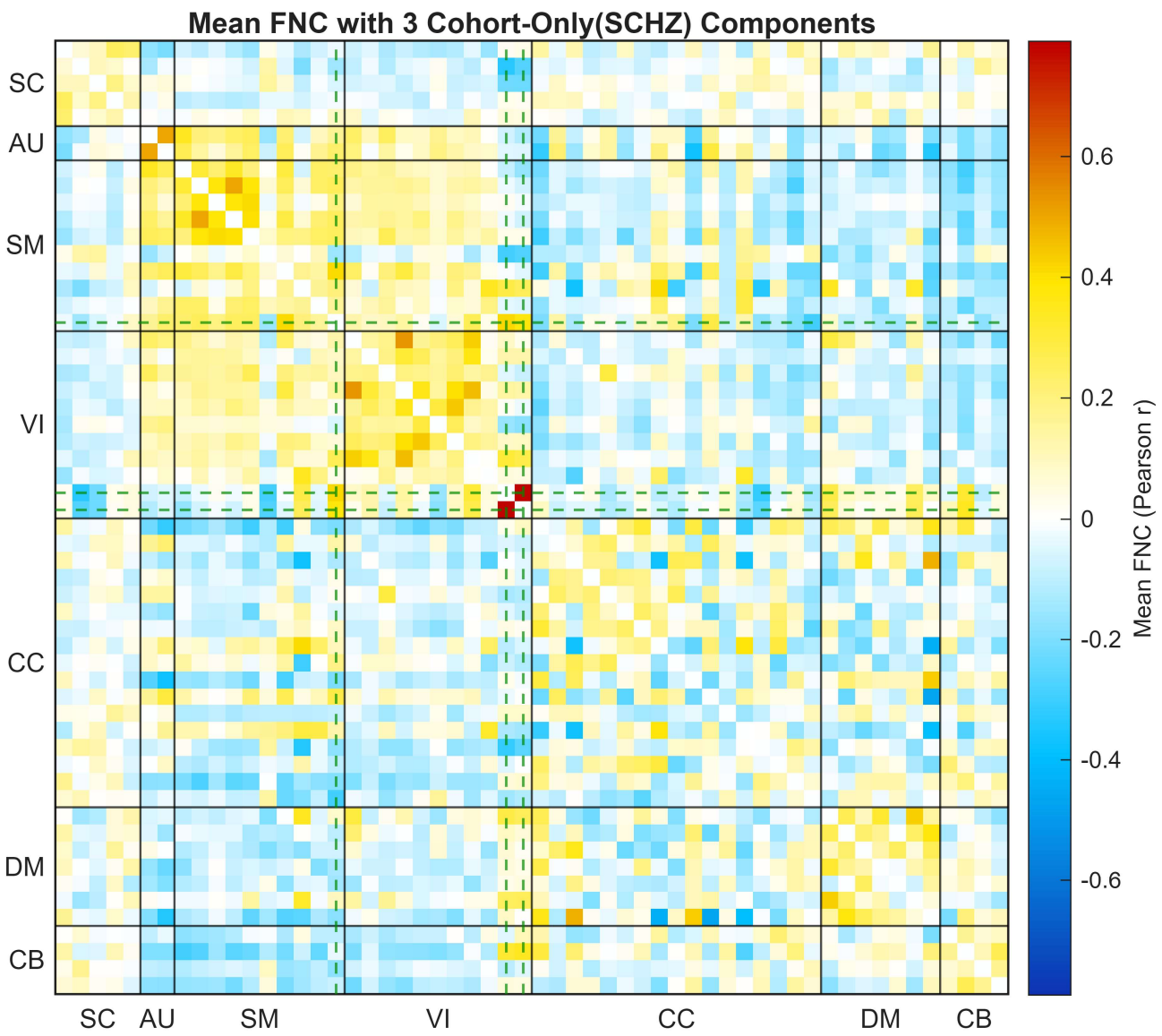

**Figure 8: FNC averaged across participants with schizophrenia (N=28), including three additional cohort-only components indicated by dashed green lines. These components integrated well into the existing modular organization of the FNC.**

## 5 CONCLUSION

We introduced CoLiG-ICA, a flexible decomposition algorithm that jointly estimates template-linked and additional free components while maximizing independence among all estimated sources. In a schizophrenia-only group analysis, CoLiG-ICA identified three additional resting-state components beyond the 53 template-linked NeuroMark components, including one sensorimotor and two lateralized visual components. These components integrated well into the existing modular FNC organization. At the subject level, CoLiG-ICA preserved the large-scale FNC structure while showing significantly lower intercomponent spatial dependence and significantly reduced motion-related variance in the template-linked components compared with MOO-ICAR. These findings suggest that free components can capture cohort-specific, subject-specific, and nuisance-related variability that would otherwise remain unresolved within template-linked networks. Although no diagnostic-group differences survived FDR correction in the UCLA CNP analysis, prior FBIRN results identified more FDR-significant effects with CoLiG-ICA (Agcaoglu et al., 2025a), motivating further validation in larger clinical cohorts.

Future work will evaluate CoLiG-ICA in larger and more heterogeneous clinical cohorts, including individuals with dementia and other neurodegenerative conditions, as well as populations with focal structural pathology or substantial anatomical abnormalities that may challenge template-guided approaches. By separating cohort- and subject-specific variability from template-linked networks, CoLiG-ICA may facilitate disease-subtype identification, biomarker discovery, and the development of individualized treatment targets.

## BIBLIOGRAPHY

Agcaoglu, O., Alacam, D., Adalı, T., Calhoun, V., Silva, R. F., Plis, S., & Bostami, B. (2024). Copula linked parallel ICA jointly estimates linked structural and functional MRI brain networks. *2024 46th Annual International Conference of the IEEE Engineering in Medicine and Biology Society (EMBC)*, 1–4. https://doi.org/10.1109/EMBC53108.2024.10781658

Agcaoglu, O., Silva, R., & Calhoun, V. (2025a). Copula Linked Group ICA: Capturing variable coupling between subject and group brain networks. *Organization for Human Brain Mapping (OHBM), Brisbane, Australia, June 24 - June 28, 2025*, 2526–2526. https://doi.org/10.5281/zenodo.15641972

Agcaoglu, O., Silva, R., & Calhoun, V. (2025b). Structure Guided Resting State Networks Estimation via Copula linked parallel ICA. *Organization for Human Brain Mapping (OHBM), Brisbane, Australia, June 24 - June 28, 2025*, 2708–2708. https://doi.org/10.5281/zenodo.15641972

Agcaoglu, O., Silva, R. F., Alacam, D., Plis, S., Adali, T., & Calhoun, V. (2024). *Copula-Linked Parallel ICA: A Method for Coupling Structural and Functional MRI brain Networks* (arXiv:2410.19774). arXiv. https://doi.org/10.48550/arXiv.2410.19774

Agcaoglu, O., Wilson, T. W., Wang, Y.-P., Stephen, J., & Calhoun, V. D. (2019). Resting state connectivity differences in eyes open versus eyes closed conditions. *Human Brain Mapping*, *40*(8), 2488–2498. https://doi.org/10.1002/hbm.24539

Allen, E. A., Erhardt, E. B., Damaraju, E., Gruner, W., Segall, J. M., Silva, R. F., Havlicek, M., Rachakonda, S., Fries, J., Kalyanam, R., Michael, A. M., Caprihan, A., Turner, J. A., Eichele, T., Adelsheim, S., Bryan, A. D., Bustillo, J., Clark, V. P., Feldstein Ewing, S. W., … Calhoun, V. D. (2011). A Baseline for the Multivariate Comparison of Resting-State Networks. *Frontiers in Systems Neuroscience*, *5*, 2. https://doi.org/10.3389/fnsys.2011.00002

Allen, E. A., Erhardt, E. B., Wei, Y., Eichele, T., & Calhoun, V. D. (2012). Capturing inter-subject variability with group independent component analysis of fMRI data: A simulation study. *NeuroImage*, *59*(4), 4141–4159. https://doi.org/10.1016/j.neuroimage.2011.10.010

Beckmann, C. F., & Smith, S. M. (2004). Probabilistic independent component analysis for functional magnetic resonance imaging. *IEEE Transactions on Medical Imaging*, *23*(2), 137–152. https://doi.org/10.1109/TMI.2003.822821

Bell, A. J., & Sejnowski, T. J. (1995). An information-maximization approach to blind separation and blind deconvolution. *Neural Computation*, *7*(6), 1129–1159. https://doi.org/10.1162/neco.1995.7.6.1129

Calhoun, V. D., Adali, T., Pearlson, G. D., & Pekar, J. J. (2001a). A method for making group inferences from functional MRI data using independent component analysis. *Human Brain Mapping*, *14*(3), 140–151. https://doi.org/10.1002/hbm.1048

Calhoun, V. D., Adali, T., Pearlson, G. D., & Pekar, J. J. (2001b). Spatial and temporal independent component analysis of functional MRI data containing a pair of task-related waveforms. *Human Brain Mapping*, *13*(1), 43–53. https://doi.org/10.1002/hbm.1024

Calhoun, V. D., Liu, J., & Adalı, T. (2009). A review of group ICA for fMRI data and ICA for joint inference of imaging, genetic, and ERP data. *NeuroImage*, *45*(1 Suppl), S163–S172. https://doi.org/10.1016/j.neuroimage.2008.10.057

Damoiseaux, J. S., Rombouts, S. a. R. B., Barkhof, F., Scheltens, P., Stam, C. J., Smith, S. M., & Beckmann, C. F. (2006). Consistent resting-state networks across healthy subjects. *Proceedings of the National Academy of Sciences of the United States of America*, *103*(37), 13848–13853. https://doi.org/10.1073/pnas.0601417103

Du, Y., & Fan, Y. (2013). Group information guided ICA for fMRI data analysis. *NeuroImage*, *69*, 157–197. https://doi.org/10.1016/j.neuroimage.2012.11.008

Du, Y., Fu, Z., Sui, J., Gao, S., Xing, Y., Lin, D., Salman, M., Abrol, A., Rahaman, M. A., Chen, J., Hong, L. E., Kochunov, P., Osuch, E. A., & Calhoun, V. D. (2020). NeuroMark: An automated and adaptive ICA based pipeline to identify reproducible fMRI markers of brain disorders. *NeuroImage: Clinical*, *28*, 102375. https://doi.org/10.1016/j.nicl.2020.102375

Erhardt, E. B., Rachakonda, S., Bedrick, E. J., Allen, E. A., Adali, T., & Calhoun, V. D. (2011). Comparison of multi-subject ICA methods for analysis of fMRI data. *Human Brain Mapping*, *32*(12), 2075–2095. https://doi.org/10.1002/hbm.21170

Griffanti, L., Douaud, G., Bijsterbosch, J., Evangelisti, S., Alfaro-Almagro, F., Glasser, M. F., Duff, E. P., Fitzgibbon, S., Westphal, R., Carone, D., Beckmann, C. F., & Smith, S. M. (2017). Hand classification of fMRI ICA noise components. *Neuroimage*, *154*, 188–205. https://doi.org/10.1016/j.neuroimage.2016.12.036

Hyvärinen, A., & Oja, E. (2000). Independent component analysis: Algorithms and applications. *Neural Networks: The Official Journal of the International Neural Network Society*, *13*(4–5), 411–430. https://doi.org/10.1016/s0893-6080(00)00026-5

Kim, T., Eltoft, T., & Lee, T.-W. (2006). Independent Vector Analysis: An Extension of ICA to Multivariate Components. In J. Rosca, D. Erdogmus, J. C. Príncipe, & S. Haykin (Eds.), *Independent Component Analysis and Blind Signal Separation* (pp. 165–172). Springer. https://doi.org/10.1007/11679363_21

Lee, J.-H., Lee, T.-W., Jolesz, F. A., & Yoo, S.-S. (2008). Independent Vector Analysis (IVA) for Group fMRI Processing of Subcortical Area. *International Journal of Imaging Systems and Technology*, *18*(1), 29–41. https://doi.org/10.1002/ima.20141

Mazziotta, J., Toga, A., Evans, A., Fox, P., Lancaster, J., Zilles, K., Woods, R., Paus, T., Simpson, G., Pike, B., Holmes, C., Collins, L., Thompson, P., MacDonald, D., Iacoboni, M., Schormann, T., Amunts, K., Palomero-Gallagher, N., Geyer, S., … Mazoyer, B. (2001). A probabilistic atlas and reference system for the human brain: International Consortium for Brain Mapping (ICBM). *Philosophical Transactions of the Royal Society of London. Series B, Biological Sciences*, *356*(1412), 1293–1322. https://doi.org/10.1098/rstb.2001.0915

McKeown, M. J., & Sejnowski, T. J. (1998). Independent component analysis of fMRI data: Examining the assumptions. *Human Brain Mapping*, *6*(5–6), 368–372.

Michael, A. M., Anderson, M., Miller, R. L., Adalı, T., & Calhoun, V. D. (2014). Preserving subject variability in group fMRI analysis: Performance evaluation of GICA vs. IVA. *Frontiers in Systems Neuroscience*, *8*, 106. https://doi.org/10.3389/fnsys.2014.00106

Poldrack, R. A., Congdon, E., Triplett, W., Gorgolewski, K. J., Karlsgodt, K. H., Mumford, J. A., Sabb, F. W., Freimer, N. B., London, E. D., Cannon, T. D., & Bilder, R. M. (2016). A phenome-wide examination of neural and cognitive function. *Scientific Data*, *3*(1), 160110. https://doi.org/10.1038/sdata.2016.110

Rashid, B., Blanken, L. M. E., Muetzel, R. L., Miller, R., Damaraju, E., Arbabshirani, M. R., Erhardt, E. B., Verhulst, F. C., van der Lugt, A., Jaddoe, V. W. V., Tiemeier, H., White, T., & Calhoun, V. (2018). Connectivity dynamics in typical development and its relationship to autistic traits and autism spectrum disorder. *Human Brain Mapping*, *39*(8), 3127–3142. https://doi.org/10.1002/hbm.24064

Sklar, A. (1959). Fonctions de répartition à n dimensions et leurs marges. *Publications de l'Institut de Statistique de l'Université de Paris*, *8*, 229–231.

Varoquaux, G., Sadaghiani, S., Pinel, P., Kleinschmidt, A., Poline, J. B., & Thirion, B. (2010). A group model for stable multi-subject ICA on fMRI datasets. *NeuroImage*, *51*(1), 288–299. https://doi.org/10.1016/j.neuroimage.2010.02.010

Vu, T., Laport, F., Yang, H., Calhoun, V. D., & Adal, T. (2024). Constrained Independent Vector Analysis With Reference for Multi-Subject fMRI Analysis. *IEEE Transactions on Bio-Medical Engineering*, *71*(12), 3531–3542. https://doi.org/10.1109/TBME.2024.3432273